\documentclass[journal]{IEEEtran}
\usepackage{amsmath,amssymb,amsthm,booktabs,multirow,graphicx,url}
\usepackage{cite}
\usepackage{algorithm}
\usepackage{algpseudocode}
\usepackage{placeins}
\usepackage{dblfloatfix}
\usepackage{flushend}
\usepackage[hidelinks]{hyperref}
\newcommand{\pos}[1]{[#1]_+}

\begin{document}
\bstctlcite{LeCo:BSTcontrol}
\title{Learning to Leverage Compliance: A Policy--Admittance Learning Framework for Robotic Insertion}
\author{Chongren~Wang\textsuperscript{*}, Minghe~Li\textsuperscript{*}, Honghua~Dai, Zhicheng~Lin, Shiyang~Wei, and Xiaokui~Yue
\thanks{\textsuperscript{*}Equal contribution.}
\thanks{Corresponding author: Honghua Dai (e-mail: hhdai@nwpu.edu.cn).}
\thanks{All authors are with the School of Astronautics and the National Key Laboratory of Aerospace Flight Dynamics, Northwestern Polytechnical University.}
\thanks{This work has been submitted to the IEEE for possible publication. Copyright may be transferred without notice, after which this version may no longer be accessible.}}
\maketitle

\begin{abstract}
Policy learning and compliant control offer a promising route to reliable autonomous assembly under pose errors and contact uncertainty.
However, combining them does not ensure coordination: the policy may continue pushing against contact while the controller yields, producing sustained loading with limited progress.
To address this problem, we propose LeCo (Leverage Compliance), a policy–admittance learning framework that trains a visual policy to leverage fixed admittance for effective insertion with reduced contact loads. A multirate feedback mechanism aggregates high-rate contact-interaction records into policy-transition rewards. An integrated conflict cost then characterizes sustained policy-loading/controller-unloading opposition, while a directional high-force tail cost captures continued-loading events within a transition. Combined with a task-completion reward, these costs encourage the policy to leverage compliance with less unproductive loading. We evaluate LeCo on four real connector-assembly tasks, obtaining an aggregate success rate of 94\%. Across tasks, mean successful-trial resultant-force and torque peaks decrease by approximately 30\% and 64\% relative to the comparison baseline. Reward ablation further shows that adding conflict shaping reduces median successful-trial contact-conditioned conflict density by approximately 53\%. These results support integrating multirate policy--admittance execution with interaction-based reward shaping to achieve effective, lower-load insertion.
Project website: \url{https://rtnow.github.io/leco/}.
\end{abstract}


\section{Introduction}
Reliable autonomous assembly requires a robot to advance a part toward its seat while correcting misalignment under pose errors and contact uncertainty. Excessive resistance prevents progress, whereas insufficient engagement prevents completion, making task motion and compliant interaction jointly important \cite{abu2020variable,suomalainen2022survey}. Visuomotor learning and reinforcement learning (RL) reduce the need to prescribe reference motions \cite{levine2016e2e,inoue2017assembly}. HIL-SERL and ConRFT further support real-robot policy acquisition through demonstrations and online corrections \cite{luo2025hilserl,chen2025conrft}.

\begin{figure}[!tbp]
\centering
\includegraphics[width=1.0\columnwidth]{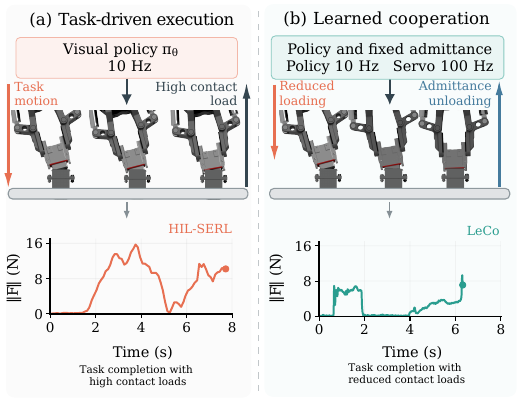}
\caption{Motivation for learning to leverage compliance. (a) Task-driven execution can complete insertion while sustaining substantial contact loads. (b) LeCo coordinates task motion with admittance unloading to reduce contact loading while preserving task completion.}
\label{fig:motivation}
\end{figure}

Compliant control adjusts execution using force feedback to respond to changing contact. Hierarchical motion--force skill reproduction \cite{zeng2025hierarchical}, learning variable compliance from demonstrations \cite{su2026admittance}, and visual--force residual RL \cite{li2026residual} combine task learning and contact regulation in different ways. Slow--fast visual--tactile policies further use feedback at different update rates to improve responsiveness \cite{xue2025reactive}. These advances provide adaptive motion and compliant execution for autonomous assembly. Their coordination within the same contact process determines how these capabilities translate into productive progress.

However, combining a learned policy with compliant control does not automatically ensure cooperation. When a connector meets an edge, the policy may continue pushing while the controller yields, producing sustained loading with limited progress. Task completion alone does not distinguish brief, necessary correction from repeated command opposition. High force can also occur during either continued loading or withdrawal from a loaded contact. Learning feedback must therefore relate contact loads to both command sources, identify sustained opposition, and preserve adjustments that restore progress. Figure~\ref{fig:motivation} illustrates contrasting contact-load profiles that motivate this objective.

The problem also concerns how interaction information is organized over time. Policy decisions and contact control typically update at different rates, so a policy transition may contain several control responses. The final observation can miss earlier opposition; an interval average can dilute a brief loading event, while a maximum alone does not describe duration. High-rate execution records must therefore be aggregated over a common transition interval, while distinguishing sustained opposition from brief high-force loading. This motivates our question: \emph{how can contact interaction across different update rates guide a policy to leverage compliance while completing its task?}

To address this question, we propose LeCo (Leverage Compliance), a policy--admittance learning framework that guides a visual policy through execution-time interaction under fixed admittance. Specifically, we aggregate high-rate contact-interaction records over transition-associated intervals to provide policy-rate learning feedback. We then construct an interaction reward from the separate policy and admittance commands. The reward combines an integrated conflict cost for simultaneous policy loading and admittance unloading with a directional tail cost for the strongest continued-loading event under high lateral force. The two terms provide complementary feedback on sustained command opposition and brief high-force loading within each recorded interval.
Combined with task completion, these costs guide the policy to exploit compliance while reducing unproductive loading. Our main contributions are:
\begin{itemize}
\item A multirate policy--admittance learning framework that connects high-rate contact feedback to policy-rate learning through interval-based aggregation under fixed compliance.
\item An interaction reward combining an integrated conflict cost with a directional tail cost to guide policy–controller cooperation through complementary feedback on sustained command opposition and brief high-force loading.
\item Validation on four real connector-assembly tasks, achieving 94\% aggregate success and lower mean successful-trial force and torque peaks across tasks. A fixed-controller reward ablation further demonstrates the complementary roles of the two costs.
\end{itemize}

\section{Related Work}
\subsection{Learning Assembly Policies}
Assembly policies must handle uncertain geometry and acquire effective contact sequences. Geometric modeling \cite{li2024multipeg} and simulation transfer \cite{tang2023industreal,tang2024automate} address these challenges through task structure and training variation; FORGE additionally conditions behavior on allowable force \cite{noseworthy2025forge}. On real robots, demonstrations guide sparse exploration \cite{nair2018overcoming}, while SAC and mixed offline--online replay provide learning foundations \cite{haarnoja2018sac,ball2023rlpd}. SERL and HIL-SERL combine these foundations with robot interaction and human corrections \cite{lee2024serl,luo2025hilserl}. Diffusion Policy and ConRFT improve action modeling and policy fine-tuning, respectively \cite{chi2023diffusion,chen2025conrft}. LeCo addresses the contact objective within such policy learning: a successful reference should also cooperate with the compliant response encountered during execution.

\subsection{Learning Compliant Contact Behavior}
The learned quantity is a key distinction among compliant-manipulation methods. Variable-compliance approaches adjust the mechanical response \cite{bogdanovic2020variable,jin2025stable}, while force-action and hybrid-controller methods learn how to command or select contact behaviors \cite{beltran2020force,liang2023preconditions}. Recent assembly studies combine residual RL with impedance control \cite{li2026residual}, learn variable impedance from demonstrations \cite{su2026admittance}, or guide compliant execution through hierarchical RL \cite{jiao2026hierarchical}. Hierarchical skill reproduction \cite{zeng2025hierarchical}, online admittance adaptation \cite{jiang2025visualimpedance}, and impact-contact control \cite{li2025implicit} further improve motion--force coordination. LeCo keeps the admittance parameters fixed and shapes the policy reference through its measured interaction with the resulting compliant response.

Force-aware imitation obtains contact supervision from demonstrations: FACTR promotes attention to force \cite{liu2025factr}, ForceMimic learns from force--motion capture \cite{liu2025forcemimic}, and Comp-ACT and UMI-FT learn compliance-related outputs \cite{kamijo2024compact,choi2026umift}. LeCo uses the policy's loading direction and the controller's unloading response to evaluate interaction generated during task learning.

\subsection{Reference Adaptation and Multirate Feedback}
Residual control learns corrections to an existing controller or reference \cite{johannink2019residual,davchev2022residual,kulkarni2022assembly}, and Reactive Diffusion Policy combines slow visual decisions with fast tactile feedback \cite{xue2025reactive}. LeCo retains this separation of task decisions and rapid contact response, but uses fast-loop records as reward supervision: integration represents sustained opposition, and directional maximization retains short loading events. 

Constraint-based and Lyapunov approaches formulate safety through explicit constrained objectives \cite{achiam2017cpo,chow2021safe}. LeCo instead shapes a task objective toward lower interaction costs; it does not impose hard force constraints. Its directional and temporal structure specifies the contact behavior being discouraged.

\begin{figure*}[!t]
\centering
\includegraphics[width=0.98\textwidth]{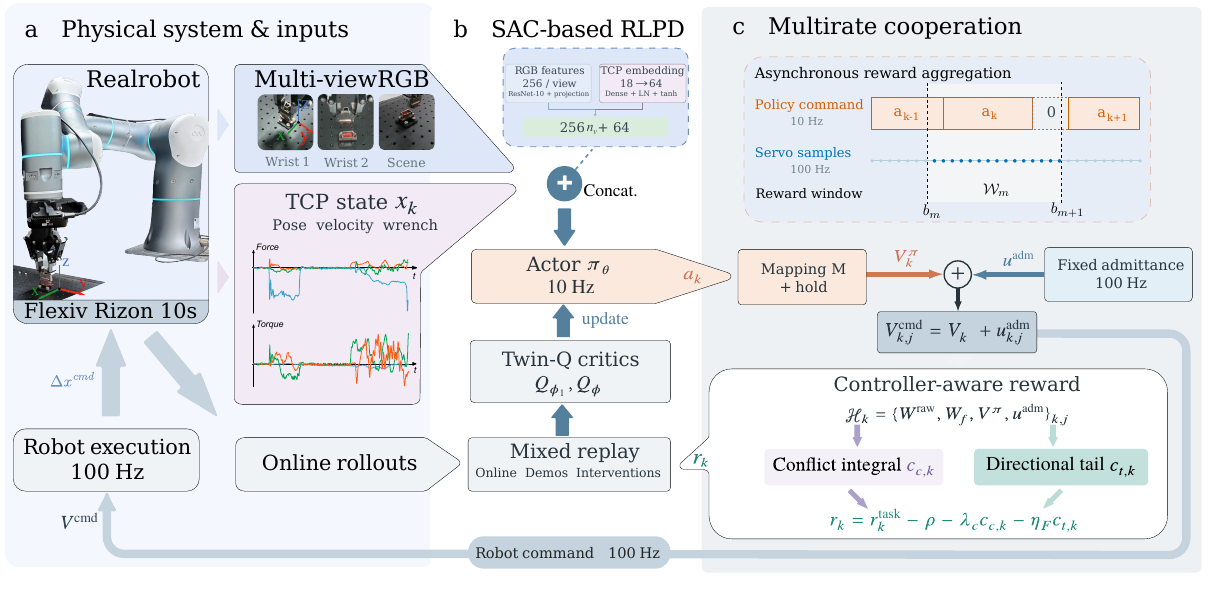}
\caption{LeCo architecture: a 10-Hz visual policy and fixed 100-Hz admittance execute task motion; recorded wrench and separate commands supply conflict-integral and directional-tail rewards to SAC/RLPD replay. In the timing schematic, $\mathcal W_m=[b_m,b_{m+1})$ is a recording window; $b_m$ and $b_{m+1}$ are the timestamps of its starting and ending controller-buffer snapshots, respectively. Rates are nominal; 0 denotes a zero policy reference while admittance continues. The command decomposition precedes limiting.}
\label{fig:overview}
\end{figure*}

\section{Problem Formulation and Methods}
LeCo connects policy-generated task motion and fixed-admittance contact response through a shared learning objective. The visual policy selects actions from observations, admittance continuously adjusts execution using wrench feedback, and the reward guides policy updates using their interaction during contact. As shown in Fig.~\ref{fig:overview}, high-rate records within each policy transition are reduced to conflict and tail costs, encouraging insertion completion with less sustained policy--admittance opposition and less continued policy loading under high lateral force.

\subsection{Problem Formulation and System Overview}
Assuming synchronous timing, we formulate the policy–admittance interaction as a partially observable Markov decision process under fixed-admittance closed-loop dynamics. Let $s_k=s(t_k)$ denote the state at decision time $t_k$, including robot, contact, and controller states. The policy selects action $a_k$ from observation $o_k$. After execution, the robot and environment evolve together with the admittance response; the conditional distribution of the next state is
\begin{equation}
p_{\mathcal A}(s_{k+1}\mid s_k,a_k).
\end{equation}
The subscript $\mathcal A$ indicates that the state transition includes the fixed admittance response. Because the observation does not fully expose contact and admittance states, both the policy and critic take the available $o_k$ as input.

The observation combines multiview images and robot state:
\begin{equation}
o_k=\{I_k^{(1)},\ldots,I_k^{(n_v)},x_k\},
\end{equation}
where $n_v$ is the number of camera views, $I_k^{(i)}$ is the RGB image from view $i$, and $x_k\in\mathbb R^{18}$ contains reset-relative TCP pose, TCP velocity, force, and torque. The action $a_k\in[-1,1]^6$ specifies Cartesian motion increments in the order $[x,y,z,r_x,r_y,r_z]$; translation and rotation are scaled by 5 mm and 0.03 rad per step. The learning objective combines task completion with reduced sustained policy--admittance conflict and transient high-force loading.

Define the common transition interval $\mathcal I_k=[t_k,t_{k+1})$. Under this timing assumption, $t_{k+1}=t_k+T_\pi$ and $o_k,o_{k+1}$ correspond to its boundaries, while $a_k$ generates the reference motion within it. State evolution and both interaction costs use this same interval. The configured policy and admittance periods are $T_\pi=0.1$ s and $\Delta t=0.01$ s; an uninterrupted synchronous interval contains ten admittance updates. More generally, $N_k$ denotes the available samples, with $t_{k,j}\in\mathcal I_k$ and $j=0,\ldots,N_k-1$.

Define $W_{k,j}=[F_{k,j}^{\top},\tau_{k,j}^{\top}]^{\top}$ as the measured wrench and $V_{k,j}^{\pi}=[(v_{k,j}^{\pi})^{\top},(\omega_{k,j}^{\pi})^{\top}]^{\top}$ as the policy reference twist obtained by mapping the action into linear and angular velocity components. Let $u_{k,j}^{\rm adm}$ denote the admittance residual twist. Within $\mathcal I_k$, the policy supplies the reference motion and the residual updates with each wrench sample:
\begin{equation}
\begin{aligned}
o_k&\xrightarrow{\pi_\theta}a_k\xrightarrow{\mathcal M}V_{k,j}^{\pi},\\
W_{k,j}&\xrightarrow{\mathcal A}u_{k,j}^{\rm adm},\quad
V_{k,j}^{\rm cmd}=V_{k,j}^{\pi}+u_{k,j}^{\rm adm}.
\end{aligned}
\label{eq:systemflow}
\end{equation}
Here $\mathcal M$ is the action mapping and $\mathcal A$ is the fixed admittance controller. Equation~\eqref{eq:systemflow} gives the command decomposition for lateral and rotational channels before final command limiting; the mapping and base-$z$ execution are detailed below. Interaction costs use both command components and wrench at each sample.

\subsection{Multirate Policy--Admittance Execution}
Before workspace and velocity limiting, the displacement-to-twist mapping is
\[
v^\pi=0.005\,a_k^{1:3}/T_\pi,\qquad
\omega^\pi=0.03\,a_k^{4:6}/T_\pi,
\]
in m/s and rad/s. Conversion uses the configured $T_\pi=0.1$ s, not the measured interval duration. A workspace-clipped target is converted back to an effective twist using the same divisor.

Let $W_{f,k,j}=[F_{f,k,j}^\top,\tau_{f,k,j}^\top]^\top$ be the filtered wrench. The residual twist $u^{\rm adm}=[(v^{\rm adm})^\top,(\omega^{\rm adm})^\top]^\top$ follows
\begin{equation}
M\dot u^{\rm adm}+Du^{\rm adm}=S\,\mathcal{D}(W_f),
\label{eq:admittance}
\end{equation}
where $M,D,S\in\mathbb{R}^{6\times6}$ are fixed diagonal virtual-inertia, damping, and axis-selection matrices, respectively, and $\mathcal{D}(\cdot)$ is an axis-wise deadband. Because all diagonal entries are constant within a sample, the implementation uses the exact zero-order-hold discretization
\begin{equation}
\begin{aligned}
\widehat u_{k,j}^{\rm adm}={}&e^{-DM^{-1}\Delta t}u_{k,j}^{\rm adm,-}\\
&+D^{-1}\!\left(I-e^{-DM^{-1}\Delta t}\right)
S\mathcal{D}(W_{f,k,j}).
\end{aligned}
\label{eq:admdisc}
\end{equation}
Here $u_{k,j}^{\rm adm,-}$ is the state before the current wrench sample, and $\widehat u_{k,j}^{\rm adm}$ is its updated candidate. Componentwise saturation gives $u_{k,j}^{\rm adm}=\operatorname{clip}(\widehat u_{k,j}^{\rm adm},-\ell,\ell)$, where $\ell$ specifies the residual-twist bounds. The previous bounded state is carried across samples and window boundaries and is zeroed only at episode reset. Execution and \eqref{eq:fu}--\eqref{eq:confpower} use this same post-update $u_{k,j}^{\rm adm}$. Lateral and rotational channels use the sum in \eqref{eq:systemflow}; the base-$z$ channel uses a held position target with an added admittance displacement.

On hardware, acquisition and inference latency cause rate fluctuations and deviations from scheduled decision boundaries. Successive timestamped controller-buffer snapshots delimit the realized recording windows; neither exactly ten samples nor exact coincidence with a single action is enforced. Costs use each sample's active command components. An action block lasts at most ten inner ticks unless superseded; afterward the policy reference twist is set to zero while admittance continues. These recorded-window rewards approximate the synchronous transition-level formulation.

Power computation uses the robot base frame and a common TCP reference point. The wrench is obtained through the Flexiv robot interface and expressed consistently with the policy reference and admittance twists. A reference-point shift satisfies
\begin{equation}
F_{\rm tip}=F_{\rm tcp},\qquad
\tau_{\rm tip}=\tau_{\rm tcp}+(p_{\rm tcp}-p_{\rm tip})\times F_{\rm tcp}.
\label{eq:wrenchshift}
\end{equation}
Here, tcp denotes the tool center point and tip a selected reference point at the connector insertion end; $p_{\rm tcp}$ and $p_{\rm tip}$ are their positions in the base frame.
For a rigid transform $T$, the dual transformations are
\begin{equation}
V_A=\operatorname{Ad}_{T}V_B,\qquad
W_A=\operatorname{Ad}_{T}^{-\top}W_B,
\end{equation}
where $\operatorname{Ad}_{T}$ is the adjoint matrix. Thus $W_A^\top V_A=W_B^\top V_B$, preserving the power inner product under consistent transformations. We use loading sign $s=-1$, with positive $sF^\top v$ and $s\tau^\top\omega$ denoting loading power.

\subsection{Multirate Analytic Cooperation Reward}
\subsubsection{Sustained policy--admittance conflict}
Across the four tasks, $P_\perp=\operatorname{diag}(1,1,0)$ and $\widetilde F=P_\perp F$ select the fixed robot-base $xy$ plane; this projection is not recomputed from the connector orientation. Thus, ``lateral'' denotes base-$xy$ force, not an independently estimated task-normal plane. The rotational cost uses all three torque channels. At inner sample $(k,j)$, define the positive parts of policy loading power and admittance unloading power:
\begin{align}
L^F_{k,j}&=\pos{s\widetilde F_{f,k,j}^\top v_{k,j}^\pi}, &
U^F_{k,j}&=\pos{-s\widetilde F_{f,k,j}^\top v_{k,j}^{\rm adm}},\label{eq:fu}\\
L^\tau_{k,j}&=\pos{s\tau_{f,k,j}^\top\omega_{k,j}^\pi}, &
U^\tau_{k,j}&=\pos{-s\tau_{f,k,j}^\top\omega_{k,j}^{\rm adm}}.\label{eq:tu}
\end{align}
These products quantify command-level opposition between the policy reference and admittance residual; using only $V^{\rm cmd}$ obscures their cancellation. They are distinct from mechanical power based on actual TCP velocity. Define conflict as:
\begin{equation}
P^F_{c,k,j}=\min(L^F_{k,j},U^F_{k,j}),\qquad
P^\tau_{c,k,j}=\min(L^\tau_{k,j},U^\tau_{k,j}).
\label{eq:confpower}
\end{equation}

\begin{figure}[!tbp]
\centering
\includegraphics[width=0.93\columnwidth]{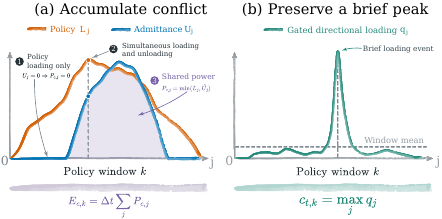}
\caption{Temporal aggregation of interaction rewards. (a) Integrating shared command power preserves conflict duration. (b) The maximum of gated loading retains brief peaks attenuated by averaging. Curves illustrate one policy window.}
\label{fig:reward_mechanism}
\end{figure}

The minimum is zero unless both components are active and attributes only their shared magnitude to conflict. 

Fig.~\ref{fig:reward_mechanism}(a) illustrates this command-level condition. Policy withdrawal, orthogonal correction, or an inactive component contributes no shared loading/unloading power.

Integrating conflict power over $\mathcal I_k$ gives quantities with energy units, without identifying actual mechanical dissipation or wasted work:
\begin{align}
E^F_{c,k}&=\Delta t\sum_{j=0}^{N_k-1}P^F_{c,k,j}, &
E^\tau_{c,k}&=\Delta t\sum_{j=0}^{N_k-1}P^\tau_{c,k,j},\\
c_{c,k}&=\frac{E^F_{c,k}}{E_F^{\rm ref}}+
\frac{E^\tau_{c,k}}{E_\tau^{\rm ref}},
\label{eq:confcost}
\end{align}
where $E_F^{\rm ref}=0.01$ J and $E_\tau^{\rm ref}=0.005$ J make the translational and rotational terms dimensionless. Filtering is used here because the integral is intended to represent persistent interaction rather than measurement spikes.

\subsubsection{Directional high-force tail cost}
The conflict integral in \eqref{eq:confcost} can underrepresent brief loading and requires admittance unloading. We therefore add a complementary translational tail term. First, raw lateral force magnitude opens a continuously differentiable smoothstep gate:
\begin{align}
\xi_{k,j}&=\operatorname{clip}\left(
\frac{\|\widetilde F^{\rm raw}_{k,j}\|-F_{\rm soft}}
{F_{\rm hard}-F_{\rm soft}},0,1\right),\\
g^F_{k,j}&=\xi_{k,j}^2(3-2\xi_{k,j}).
\label{eq:gate}
\end{align}
Thus $g^F=0$ below $F_{\rm soft}$, changes smoothly between the two thresholds, and equals one above $F_{\rm hard}$. We set $F_{\rm soft}=2$ N and $F_{\rm hard}=3$ N. Raw rather than filtered force is used so a fast contact transient is retained.

The tail term targets continued policy loading under high lateral force. To make this term responsive to loading direction, we use the positive part of the raw policy-loading power:
\begin{equation}
P^{\rm raw}_{L,k,j}=\pos{s(\widetilde F^{\rm raw}_{k,j})^\top v_{k,j}^\pi}.
\label{eq:rawload}
\end{equation}
When the policy unloads or moves orthogonally to the force, \eqref{eq:rawload} is zero even if the force gate is open. Over the same interval $\mathcal I_k$, the per-transition tail cost is
\begin{equation}
c_{t,k}=\max_{0\le j<N_k}
\left[g^F_{k,j}\min\left(\frac{P^{\rm raw}_{L,k,j}}
{P_F^{\rm ref}},q_{\max}\right)\right],
\label{eq:tail}
\end{equation}
with $P_F^{\rm ref}=0.5$ W and $q_{\max}=10$. In contrast to the conflict integral, the sample maximum approximates the supremum of gated loading over $\mathcal I_k$. Thus $0\le c_{t,k}\le q_{\max}$; with $\eta_F=0.1$, the magnitude of the tail penalty in one reward is bounded by $\eta_Fq_{\max}=1.0$.

Fig.~\ref{fig:reward_mechanism} summarizes both the physical tests and their complementary temporal reductions. Conflict requires simultaneous loading and unloading and integrates their filtered shared power, thereby retaining duration. The tail term first gates raw high force, then retains only positive policy loading and takes a maximum, thereby preserving a brief worst sample without penalizing policy unloading or orthogonal motion.

\begin{algorithm}[!tbp]
\caption{LeCo: Synchronous Transition-Level Formulation}
\label{alg:posttraining}
\small
\begin{algorithmic}[1]
\Require initial parameters $\theta_0$; fixed admittance $\mathcal A$;
demonstration replay $\mathcal D_E$; $\lambda_c,\eta_F$
\State $\theta\gets\theta_0$; initialize empty online replay $\mathcal D_O$
\For{each episode}
  \State reset environment and admittance state; observe $o_k$
  \While{episode is not terminal}
    \State at $t_k$, sample $a_k\sim\pi_\theta(\cdot\mid o_k)$; replace it during intervention; set the reference for $\mathcal I_k$
    \State $E^F_{c,k},E^\tau_{c,k},c_{t,k}\gets0$; retain admittance state
    \For{$j=0,\ldots,N_k-1$} \Comment{samples in $\mathcal I_k$}
      \State acquire $W^{\rm raw}_{k,j}$ and filtered $W_{f,k,j}$
      \State update $u^{\rm adm,-}_{k,j}$ by \eqref{eq:admdisc}; saturate to obtain $u^{\rm adm}_{k,j}$
      \State execute the policy reference with fixed admittance and motion limits
      \State compute $P^F_{c,k,j},P^\tau_{c,k,j}$ using \eqref{eq:fu}--\eqref{eq:confpower}
      \State $E^F_{c,k}\mathrel{+}=\Delta tP^F_{c,k,j}$;
      $E^\tau_{c,k}\mathrel{+}=\Delta tP^\tau_{c,k,j}$
      \State compute $q_{k,j}$, the bracketed term in \eqref{eq:tail}
      \State $c_{t,k}\gets\max(c_{t,k},q_{k,j})$
    \EndFor
    \State at $t_{k+1}$, observe $o_{k+1}$; obtain $r_k^{\rm task}$ from the success classifier and $d_k$ from success or the 150-step limit; compute $c_{c,k}$ by \eqref{eq:confcost}
    \State $r_k\gets r_k^{\rm task}-\rho-\lambda_cc_{c,k}-\eta_Fc_{t,k}$
    \State store $(o_k,a_k,r_k,o_{k+1},d_k)$ in $\mathcal D_O$; copy corrections to $\mathcal D_E$
    \If{$|\mathcal D_O|\ge100$}
      \State sample 128 transitions each from $\mathcal D_O$ and $\mathcal D_E$
      \State perform two critic updates using \eqref{eq:target}
      \State update actor, temperature, and target networks
    \EndIf
    \State $o_k\gets o_{k+1}$
  \EndWhile
\EndFor
\end{algorithmic}
\end{algorithm}

\subsection{Controller-Aware Learning}
The final reward stored at policy rate is
\begin{equation}
\begin{aligned}
r_k&=r_k^{\rm task}-\rho-\lambda_cc_{c,k}-\eta_Fc_{t,k},\\
\rho&=0.01,\qquad\lambda_c,\eta_F\ge0.
\end{aligned}
\label{eq:reward}
\end{equation}
Following HIL-SERL~\cite{luo2025hilserl}, a trained binary image classifier gives $r_k^{\rm task}=1$ for completed assembly and $0$ otherwise. Success terminates the episode; otherwise, reaching $H=150$ policy steps terminates it as a timeout in all four tasks. The time penalty and the two interaction costs in \eqref{eq:reward} are the remaining reward terms. We set the conflict weight to $\lambda_c=0.025$ and the tail weight to $\eta_F=0.1$. For a next action $a'\sim\pi_\theta(\cdot|o_{k+1})$, the target is
\begin{equation}
\begin{aligned}
y_k={}&r_k+\gamma(1-d_k)\min_iQ_{\bar\phi_i}(o_{k+1},a'),\\
&i\in\{1,2\}.
\end{aligned}
\label{eq:target}
\end{equation}
where $d_k$ is the terminal indicator and $\gamma=0.97$. 

Algorithm~\ref{alg:posttraining} presents the nominal transition-level formulation, rather than the scheduling of asynchronous hardware threads. Each episode resets the environment and admittance state. At $t_k$, the policy selects $a_k$ from $o_k$, replaced by the operator-corrected action during intervention. Within $\mathcal I_k$, the reference and high-rate admittance jointly drive the robot. Admittance remains continuous across intervals; only the reward accumulators are reset.

Both costs summarize the same $\mathcal I_k$: conflict integrates shared loading/unloading power, whereas the tail term retains the largest gated policy-loading sample. At $t_{k+1}$, these costs, the task reward, and the next observation define the stored transition through \eqref{eq:reward}. The reward characterizes the joint closed-loop evolution from $s_k$, not the isolated causal effect of $a_k$; earlier contact and controller memory are part of the evolving state. Realized recording windows follow the implementation convention described above.

The complete transition $(o_k,a_k,r_k,o_{k+1},d_k)$ is stored in online replay $\mathcal D_O$, with $a_k$ denoting the action actually executed. Corrected transitions are also added to demonstration/intervention replay $\mathcal D_E$. The learner samples mixed batches from both buffers, updates the critics using \eqref{eq:target}, and then updates the policy, entropy temperature, and target networks. Online experience captures the current policy's contact behavior, while demonstrations and corrections provide examples of task completion and contact adjustment, allowing learning to use both autonomous exploration and human guidance.

\section{Experiments}
To evaluate the proposed LeCo in real contact-rich assembly, we conduct experiments on four connector-assembly tasks, focusing on three questions. \emph{RQ1:} Compared with existing learning methods, can LeCo reduce contact loads during training and autonomous execution while improving task success and execution efficiency? \emph{RQ2:} Can rewards constructed from policy--admittance interaction reduce sustained opposition while maintaining productive contact? \emph{RQ3:} How do the conflict and directional tail costs affect task acquisition and contact quality?

We first introduce the tasks, hardware, baselines, and evaluation metrics, then analyze the four-task comparisons, and finally address RQ2 and RQ3 through reward ablations.

\subsection{Experimental Setup}
\subsubsection{Tasks and platform}
\begin{figure}[!tbp]
\centering
\includegraphics[width=\columnwidth]{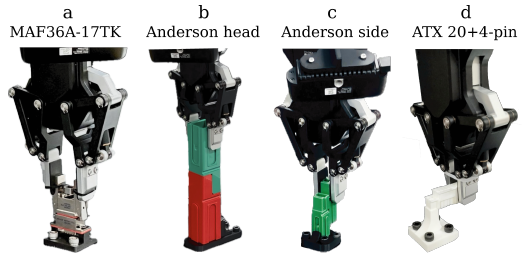}
\caption{Four real connector-assembly tasks: (a) aviation-connector insertion; (b) Anderson-head insertion; (c) Anderson-side interlocking; (d) ATX 20+4-pin connector housing assembly.}
\label{fig:task_suite}
\end{figure}

As shown in Fig.~\ref{fig:task_suite}, the four real assembly tasks differ in connector geometry, mating constraints, and contact modes. \emph{MAF36A-17TK aviation-connector insertion} requires early pose correction to avoid jamming and sufficient axial force for final seating. \emph{Anderson-head insertion} requires maintained alignment after housing contact. \emph{Anderson-side interlocking} joins lateral dovetail features, where small offsets can create substantial resistance. \emph{ATX 20+4-pin connector housing assembly} requires aligning the mating guide features of the 20-pin and 4-pin housings of a motherboard power connector and sliding them together to complete the mechanical assembly, demanding precise local alignment. Together, the tasks test the balance between necessary advancement and unproductive loading.

The experiments use a Flexiv Rizon 10s robot arm, with demonstrations and training corrections provided through a SpaceMouse. Visual observations come from Orbbec Gemini 335 wrist cameras and a WHEELTEC C100 scene camera. The aviation and ATX 20+4-pin tasks use two wrist views and one scene view; both Anderson tasks use two wrist views. A Jetson AGX runs execution and an RTX 4090 server runs learning. Each task uses 40 successful demonstrations. LeCo uses a ResNet-10 encoder with $128\times128$ images.

For LeCo, the policy and admittance rates are configured at 10/100 Hz, with a 15-Hz wrench-filter cutoff. The admittance controller enables all six axes ($S=I_6$). Table~\ref{tab:admpar} reports the fixed virtual inertia, damping, deadbands, and residual-twist bounds $\ell$; these parameters remain unchanged during learning and evaluation.

\begin{table}[!tbp]
\caption{Fixed admittance parameters in axis order $x,y,z,r_x,r_y,r_z$. Translational/rotational units: $M$, kg / N\,m\,s$^2$/rad; $D$, N\,s/m / N\,m\,s/rad; Deadband, N / N\,m; Limit, m/s / rad/s (residual-twist bounds).}
\label{tab:admpar}
\centering\small
\setlength{\tabcolsep}{5pt}
\begin{tabular}{lrrrrrr}
\toprule
 & $x$ & $y$ & $z$ & $r_x$ & $r_y$ & $r_z$\\
\midrule
$M$ &40&40&100&0.375&0.375&0.75\\
$D$ &800&800&2000&7.5&7.5&15\\
Deadband &0.5&0.5&1.0&0.03&0.03&0.03\\
Limit &0.010&0.010&0.020&0.20&0.20&0.10\\
\bottomrule
\end{tabular}
\end{table}

\subsubsection{Baselines and metrics}
\emph{HIL-SERL}~\cite{luo2025hilserl} learns from demonstrations, sparse rewards, and online corrections. \emph{ConRFT}~\cite{chen2025conrft} combines offline training with online reinforced fine-tuning; we retain consistency-policy learning with a ResNet-10 encoder.
All methods share the low-level robot controller, asynchronous actor--learner setup, task-specific demonstration counts, visual layouts, action spaces, reset-seed sets, and operator. HIL-SERL and ConRFT use no additional admittance layer; LeCo adds the fixed admittance in Table~\ref{tab:admpar}. Policy structures and objectives remain method specific.

We measure \emph{contact load} by per-episode lateral-force peaks $F_{xy}=\sqrt{F_x^2+F_y^2}$ from the robot's wrench interface, including autonomous, assisted, and failed training episodes; \emph{training autonomous success} by the fraction of the last ten episodes completed without human motion correction; and \emph{efficiency} by intervention steps, time to a specified success level, and successful evaluation length. Frozen-policy evaluation uses deterministic actions and the same classifier-based success criterion as HIL-SERL~\cite{luo2025hilserl}.

Training statistics use the last ten complete episodes, with an expanding prefix before ten are available. Wrench peaks use window medians, and intervention ratios are step weighted. Online time excludes ConRFT's offline training.

\begin{figure*}[!t]
\centering
\includegraphics[width=0.94\textwidth,trim={0 10 0 8},clip]{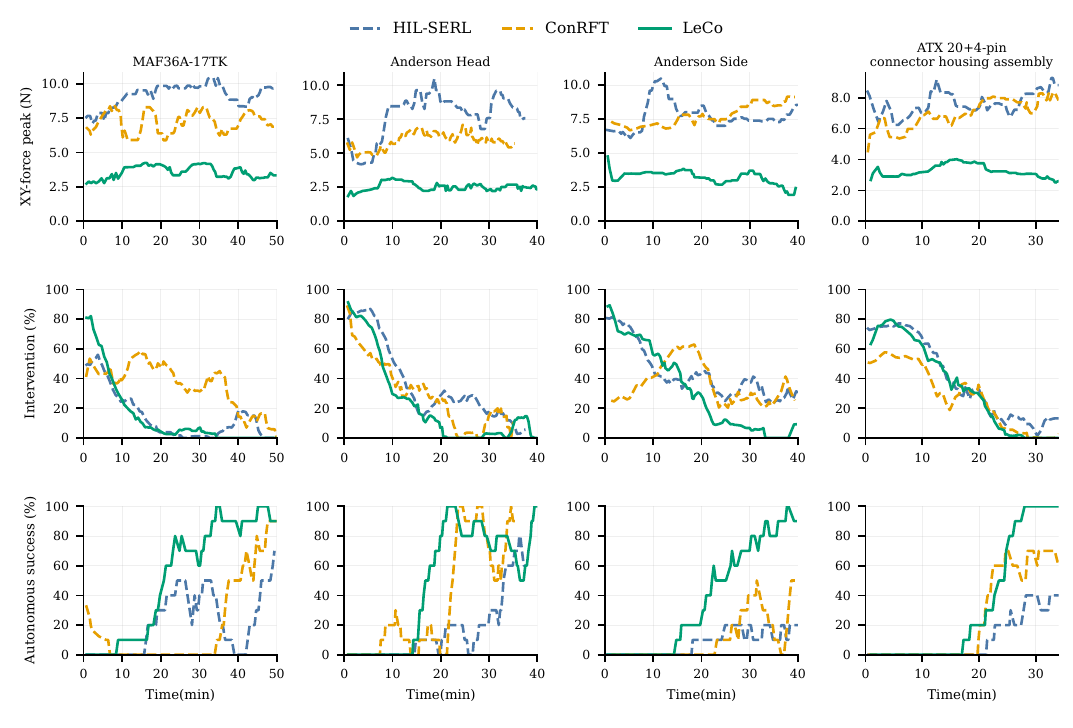}
\caption{Online learning on four tasks. Rows show episode-wise XY-force peaks in the recorded frame, intervention ratios, and autonomous success. Force curves are medians of up to ten recent episode peaks. Online time excludes ConRFT offline training.}
\label{fig:multitask_training}
\end{figure*}

\subsection{Comparison With Learning Baselines}
\subsubsection{Training contact loads and learning behavior}
To address RQ1, we first examine the contact loads experienced during learning and then autonomous task performance. The three rows of Fig.~\ref{fig:multitask_training} show force peaks, intervention ratios, and autonomous success for all four tasks.

Within the displayed training windows, LeCo exhibits generally lower lateral-force peak curves across the four tasks. For Anderson-side interlocking, the final displayed complete ten-episode window has a median lateral-force peak of 2.43 N for LeCo, compared with 9.12 N for ConRFT and 8.54 N for HIL-SERL, approximately 73\% lower than ConRFT. These statistics include assisted and failed episodes and characterize the contact loads experienced during the displayed portions of learning.

Beyond lower contact loads, LeCo also acquires autonomous task behavior faster. On aviation-connector insertion, our method first reaches an 80\% ten-episode autonomous-success rate at 23.7 min, versus 46.0 min for ConRFT, a reduction of approximately 48\% in online time to this level. HIL-SERL does not reach that level in the recorded training. Our method also reaches it first on Anderson-head and ATX 20+4-pin housing assembly. On Anderson-side interlocking, only LeCo reaches this threshold within the displayed training window.

Taken together, force peaks, intervention ratios, and autonomous success show that lower loads are not obtained by abandoning contact or relying on more human takeover. The policy acquires autonomous insertion earlier while reducing contact loads and requiring fewer corrections late in training. The online results for RQ1 therefore demonstrate a joint improvement in contact quality and autonomous learning efficiency, rather than a reduction in one force metric alone. The subsequent fixed-controller reward ablation examines the roles of tail and conflict shaping.

\subsubsection{Post-training autonomous performance}
After training, we fix policy parameters and evaluate autonomous execution using the same reset-seed set within each task. Table~\ref{tab:multitask_baseline} summarizes success, completion steps, and contact peaks across all four tasks. Steps are medians over successful trials; all three peak metrics are reported as means and sample standard deviations of successful-trial maxima.

\begin{table*}[!t]
\caption{Autonomous evaluation: successful-trial median steps and mean$\pm$sample SD of trialwise peaks. Resultant force/torque use three-axis norms. Bold: highest success or lowest mean/steps per task, including ties.}
\label{tab:multitask_baseline}
\centering\small
\setlength{\tabcolsep}{3pt}
\begin{tabular}{llccccc}
\toprule
Task & Method & Success & Steps & $F_{xy}$ peak (N) & Force peak (N) & Torque peak (N\,m)\\
\midrule
\multirow{3}{*}{MAF36A-17TK}
& HIL-SERL &14/20&90&$9.809\pm1.239$&$11.950\pm0.965$&$2.304\pm0.252$\\
& ConRFT &17/20&46&$6.658\pm1.833$&$10.701\pm1.693$&$1.569\pm0.433$\\
& LeCo &\textbf{19/20}&\textbf{33}&$\mathbf{3.121\pm0.669}$&$\mathbf{7.760\pm0.807}$&$\mathbf{0.722\pm0.147}$\\
\midrule
\multirow{3}{*}{Anderson head}
& HIL-SERL &26/30&58.5&$6.147\pm1.056$&$7.856\pm1.119$&$1.701\pm0.242$\\
& ConRFT &\textbf{30/30}&\textbf{45}&$5.702\pm0.862$&$7.293\pm1.409$&$1.563\pm0.249$\\
& LeCo &\textbf{30/30}&49&$\mathbf{2.236\pm0.654}$&$\mathbf{6.188\pm1.309}$&$\mathbf{0.596\pm0.172}$\\
\midrule
\multirow{3}{*}{Anderson side}
& HIL-SERL &4/20&88&$10.822\pm3.539$&$12.883\pm1.969$&$2.165\pm0.667$\\
& ConRFT &13/20&\textbf{50}&$9.531\pm0.887$&$12.437\pm1.049$&$2.089\pm0.137$\\
& LeCo &\textbf{17/20}&52&$\mathbf{2.849\pm0.651}$&$\mathbf{7.379\pm1.691}$&$\mathbf{0.640\pm0.151}$\\
\midrule
\multirow{3}{*}{ATX 20+4-pin}
& HIL-SERL &22/30&56&$7.876\pm1.015$&$9.009\pm1.041$&$1.601\pm0.178$\\
& ConRFT &18/30&69&$7.617\pm1.289$&$9.096\pm1.555$&$1.537\pm0.357$\\
& LeCo &\textbf{28/30}&\textbf{26}&$\mathbf{2.931\pm0.742}$&$\mathbf{6.454\pm0.929}$&$\mathbf{0.483\pm0.147}$\\
\bottomrule
\end{tabular}
\end{table*}

On the aviation-connector task, LeCo succeeds in 19/20 trials and reduces median successful length from ConRFT's 46 steps to 33 steps. Mean successful-trial force peaks decrease from 10.70 to 7.76 N, and torque peaks from 1.57 to 0.72 N\,m, indicating improvements in both task completion and contact quality. Anderson-head insertion exhibits a different efficiency--load relationship: both LeCo and ConRFT complete all 30 trials, and ConRFT requires four fewer median steps, but LeCo reduces mean force and torque peaks by approximately 15\% and 62\%, respectively. LeCo's benefit therefore does not require the fewest steps on every task; it reduces contact loads while maintaining task success. On the ATX 20+4-pin task, LeCo succeeds in 28/30 trials with a median of 26 steps and lower force and torque peaks than the baselines.

Table~\ref{tab:multitask_baseline} shows two complementary benefits. On aviation-connector insertion and ATX 20+4-pin housing assembly, LeCo combines higher success, fewer completion steps, and lower contact peaks, indicating that improved contact quality is not obtained by sacrificing task progress. On Anderson-head insertion and Anderson-side interlocking, LeCo requires slightly more policy steps than ConRFT in successful trials but achieves lower lateral-force and torque peaks. In particular, their equal success on Anderson-head insertion does not imply equal contact quality.

These differences reflect the joint need for advancement and unloading during assembly. Final seating of the aviation connector requires adequate axial force, whereas dovetail interlocking requires avoiding sustained loading caused by misalignment under a constrained guide. A policy must therefore be evaluated not only by whether it reduces force, but also by whether that reduction accompanies productive assembly. LeCo improves both completion efficiency and contact peaks on the aviation and ATX 20+4-pin tasks, while primarily improving the contact quality of already successful behavior on Anderson head. Its benefit thus does not require the fewest steps on every task. Together with reduced intervention dependence during learning, these results support the acquisition of autonomous assembly with less unproductive contact loading. These complete-system comparisons assess policy learning and execution jointly; the following ablation examines reward contributions under identical fixed admittance.

\subsection{Ablation}
\label{sec:fourway_ablation}
\subsubsection{Reward variants and evaluation metrics}
The $2\times2$ ablation on MAF36A separates the two cooperation costs. A0 uses the task reward alone, and A1 adds conflict shaping. A2 is a \emph{tail-only force-aware baseline} with directional high-force shaping; LeCo additionally includes the full lateral-and-rotational conflict cost. All conditions share the time penalty, observations, action space, demonstrations, and fixed admittance controller. Table~\ref{tab:fourway_task} reports the weights and task performance.

\begin{table}[!tbp]
\caption{Reward-component ablation. Steps are medians over successful trials.}
\label{tab:fourway_task}
\centering\small
\setlength{\tabcolsep}{3pt}
\begin{tabular}{lccrr}
\toprule
 & $\lambda_c$ & $\eta_F$ & Success & Steps\\
\midrule
A0 &0&0&15/20&43.0\\
A1 &0.025&0&12/20&88.5\\
A2 &0&0.1&18/20&46.0\\
LeCo &0.025&0.1&19/20&33.0\\
\bottomrule
\end{tabular}
\end{table}

\begin{figure}[!tbp]
\centering
\includegraphics[width=0.8\columnwidth,trim={0 6 0 12},clip]{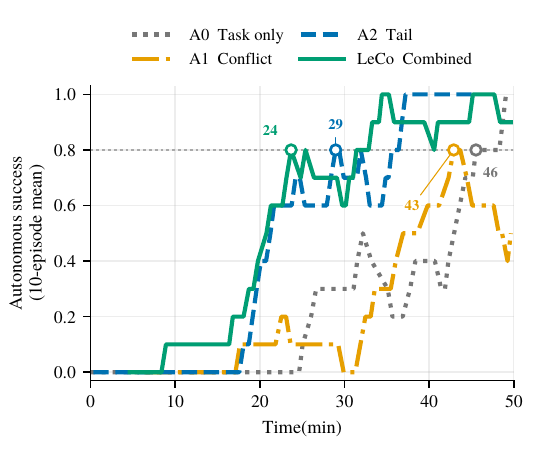}
\caption{Autonomous success in the reward ablation. Curves are ten-episode moving means; markers indicate the first window reaching 80\% autonomous success.}
\label{fig:training_ablation}
\end{figure}

\begin{figure*}[!t]
\centering
\includegraphics[width=0.94\textwidth]{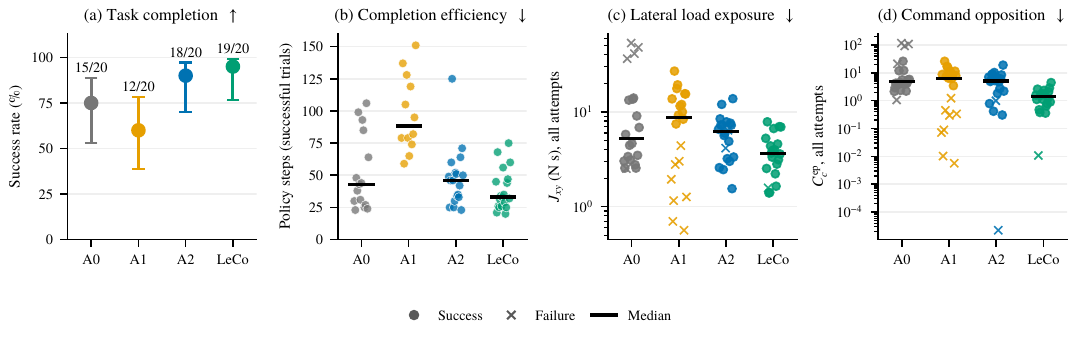}
\caption{Reward ablation: (a) success with 95\% Wilson intervals; (b) successful-trial steps; (c,d) all-attempt lateral-force exposure/conflict (log axes). Points/crosses denote successes/failures; black segments are medians.}
\label{fig:fourway_ablation}
\end{figure*}

The mechanical analysis includes every attempt because success-conditioned statistics alone can hide long-contact failures, while all-attempt peak statistics can reward policies that barely touch the task. From the raw base-frame wrench at the TCP, define $F_{xy}=\sqrt{F_x^2+F_y^2}$, $\tau=\|\boldsymbol\tau\|_2$, and
\begin{equation}
J_{xy}=\int F_{xy}(t)\,dt,\quad
J_\tau=\int\tau(t)\,dt,\quad
C_c^{\rm ep}=\sum_k c_{c,k}.
\end{equation}
The two integrals measure cumulative lateral-force and torque exposure, respectively. $C_c^{\rm ep}$ measures accumulated opposition between the policy and admittance commands.

To assess policy–admittance coordination during contact, we measure conflict per unit contact time and magnitude-weighted rotational command cancellation. A larger cancellation index indicates greater cancellation between the rotational command components.
For sample $i$, $I_i=\mathbf1_{\{\|F_i^{\rm raw}\|>1\,{\rm N}\}}$ indicates contact and $\delta_i$ is the recorded time increment. The metrics are
\begin{subequations}\label{eq:contact_metrics}
\begin{equation}
T_c=\sum_i I_i\delta_i,\quad
\ell_i=P^F_{c,i}/E_F^{\rm ref}+P^\tau_{c,i}/E_\tau^{\rm ref},\label{eq:contact_density}
\end{equation}
\begin{equation}
\bar c_c=\frac{\sum_i I_i\ell_i\delta_i}{T_c},\label{eq:contact_conflict}
\end{equation}
\begin{equation}
\chi_\omega=\frac{\sum_{i\in\mathcal V}I_i\delta_i
(b_i-\|\omega_i^\pi+\omega_i^{\rm adm}\|)}
{\sum_{i\in\mathcal V}I_i\delta_i b_i},\label{eq:rotation_cancellation}
\end{equation}
\end{subequations}
where $b_i=\|\omega_i^\pi\|+\|\omega_i^{\rm adm}\|$ and $\mathcal V$ includes finite samples with $b_i>10^{-8}$ rad/s. A zero denominator yields an undefined value, excluded from that metric rather than replaced by zero. Here $T_c$ is cumulative above-threshold contact, not elapsed time from first contact to termination. Recorded increments are clipped to $[0,0.1]$ s, with the first increment set to the median positive increment; reward accumulation instead uses configured $\Delta t$.

\subsubsection{The combined reward gives the best overall performance}
LeCo leads the task and accumulated-interaction criteria simultaneously. It achieved 19/20 deterministic successes, completed successful trials in a median of 33.0 policy steps, and produced the lowest median and P90 lateral-force exposure, torque exposure, and accumulated conflict across all attempts. These results establish the best combined completion, efficiency, and contact quality among the four reward variants.

Fig.~\ref{fig:fourway_ablation} compares task completion and contact cost together: panels (a,b) show success and successful-trial length, while (c,d) report accumulated lateral force and command conflict across all attempts, including failures. LeCo combines high completion and short successful trajectories with low all-attempt accumulated interaction, showing that reduced loading accompanies productive progress.

\begin{figure*}[!t]
\centering
\includegraphics[width=0.88\textwidth,trim={0 6 0 6},clip]{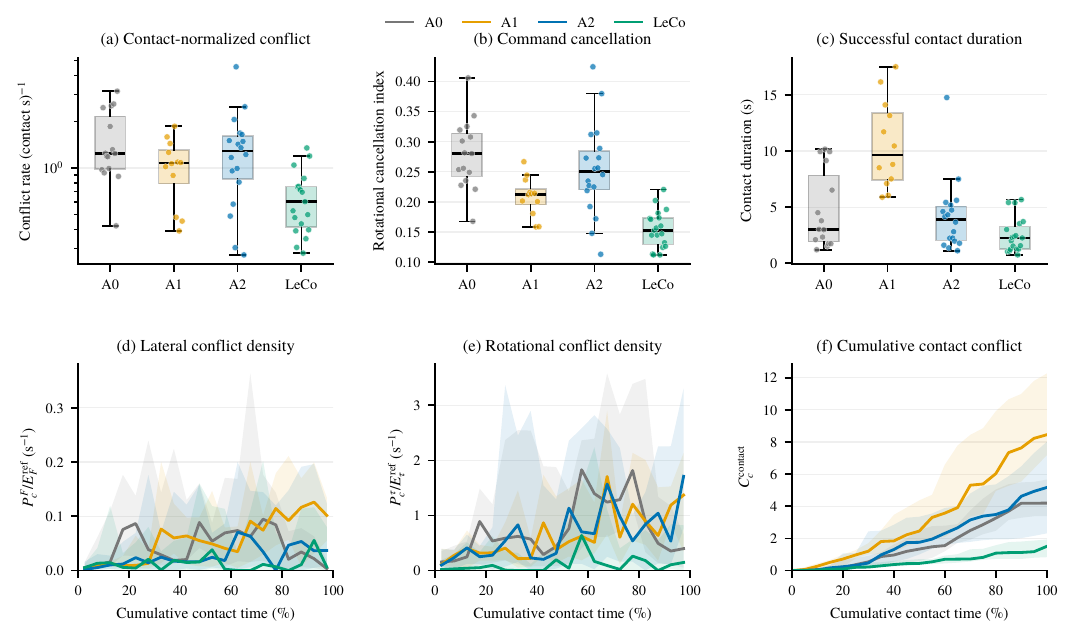}
\caption{Successful-trial 100-Hz records: (a,b) contact conflict/cancellation; (c) contact duration; (d,e) lateral/rotational conflict density averaged within 20 equal-duration contact bins; (f) cumulative contact conflict. The horizontal axis in (d)--(f) is cumulative above-threshold contact time normalized per trial. Boxes/shading: interquartile ranges; lines: medians.}
\label{fig:command_contact_mechanism}
\end{figure*}

\begin{table*}[!t]
\caption{Contact statistics across all attempts ($n=20$ per method, including failures). Entries are median / P90 across episodes. Bold indicates the lowest median and P90 in each column.}
\label{tab:fourway_load}
\centering\small
\setlength{\tabcolsep}{4pt}
\begin{tabular}{lrrrrrr}
\toprule
&\multicolumn{3}{c}{Instantaneous peaks}&\multicolumn{3}{c}{Accumulated interaction}\\
\cmidrule(lr){2-4}\cmidrule(lr){5-7}
 & $F_{xy}$ (N) & $|F_z|$ (N) & $\tau$ (N\,m)
 & $J_{xy}$ (N\,s) & $J_\tau$ (N\,m\,s) & $C_c^{\rm ep}$\\
\midrule
A0 &3.389 / 4.453&8.487 / 9.563&0.827 / 1.115&5.216 / 41.904&1.216 / 16.332&5.007 / 92.552\\
A1 &\textbf{3.028 / 3.871}&8.761 / 10.088&\textbf{0.756} / 0.981&8.815 / 17.800&1.951 / 4.040&6.320 / 15.340\\
A2 &3.414 / 4.304&\textbf{7.915 / 8.571}&0.827 / 0.988&6.170 / 8.804&1.378 / 2.049&5.027 / 9.077\\
\textbf{LeCo} &3.298 / 4.008&7.997 / 8.890&0.776 / \textbf{0.938}&\textbf{3.637 / 6.943}&\textbf{0.808 / 1.521}&\textbf{1.379 / 2.606}\\
\bottomrule
\end{tabular}
\end{table*}

\subsubsection{Complementary roles of tail and conflict shaping}
The tail term is the component most directly associated with task acquisition in this ablation. Without conflict shaping, A2 reached an 80\% autonomous-success training window at approximately 29 min, compared with 46 min for A0, and improved frozen-policy success from 15/20 to 18/20. With conflict shaping present, adding the tail term changes A1 into LeCo: the first 80\% training window moves from approximately 43 to 24 min, frozen-policy success rises from 12/20 to 19/20, and median successful length falls from 88.5 to 33.0 steps. Both the A0--A2 and A1--LeCo comparisons indicate that directional tail shaping supports task acquisition in this ablation.

The failure distribution supports the same mechanism. A2 reduced the number of failures from five to two and reduced P90 lateral exposure from 41.90 to 8.80 N\,s. Because the tail cost activates when high lateral force coincides with continued policy loading, it encourages the policy to leave unproductive loading modes while preserving corrective and unloading motion.

With directional high-force tail shaping already present, A2--LeCo tests the incremental value of the full conflict cost under identical admittance. Adding conflict shaping reduces median successful length from 46 to 33.0 steps and lowers lateral-force/torque exposure, accumulated and contact-conditioned conflict, and rotational command cancellation. The benefits therefore span task efficiency, load exposure, and measured command opposition.

Fig.~\ref{fig:command_contact_mechanism}(a)--(c) examines both command opposition and contact duration. Relative to A2, LeCo reduces successful-trial median contact-conditioned conflict density from 1.29 to 0.60 s$^{-1}$, rotational cancellation from 0.25 to 0.15, and contact duration from 3.91 to 2.26 s. Thus, shorter contact accompanies lower opposition per unit contact time, rather than merely reducing accumulated quantities by terminating earlier. Panels (d,e) show that the reduction comes mainly from rotational conflict under the reference-energy scaling, while lateral-conflict profiles largely overlap. Panel (f) confirms the cumulative benefit: LeCo's median terminal contact integral is 1.50, compared with 5.17 for A2. A1 accumulates more conflict despite lower conflict density because contact persists longer. Together, these results indicate that LeCo improves cooperation by reducing sustained opposition while maintaining efficient task progress.

\subsubsection{Accumulated interaction across complete attempts}
Table~\ref{tab:fourway_load} distinguishes instantaneous peaks from interaction over the complete trial. Peak values overlap across conditions, and LeCo is not best on every peak metric; the more consistent differences occur in accumulated exposure and conflict. Relative to A2, LeCo reduces median lateral-force exposure from 6.17 to 3.64 N\,s, torque exposure from 1.38 to 0.81 N\,m\,s, and accumulated conflict from 5.03 to 1.38. All three P90 values also decrease, indicating improvements in both typical trials and the higher-exposure tail. Conversely, although A1 reduces some instantaneous opposition measures, its longer contact duration prevents these reductions from translating automatically into lower cumulative loads.

A2--LeCo thus supports the incremental value of the complete conflict module: improved task efficiency and load exposure accompany reduced command opposition. This evidence concerns complementarity with directional tail shaping under fixed admittance; it neither ranks all force-magnitude penalties nor establishes the indispensability of individual conflict operators.

\section{Conclusion}
LeCo teaches a visual policy to leverage fixed admittance by converting execution-time interaction into reinforcement-learning rewards. A conflict integral captures sustained policy-loading/admittance-unloading opposition, while a directional tail cost retains brief high-force loading events. The controller parameters, policy interface, and deployment structure remain unchanged.

Four real connector-assembly tasks show improved autonomous acquisition and reduced contact loads. Ablations identify complementary roles: tail shaping supports reliable insertion, and conflict shaping reduces opposition within productive behavior. Their combination shortens successful trajectories and lowers accumulated force, torque exposure, and command conflict, without implying lower instantaneous loads at every moment. These findings support using an existing compliant controller as a source of learning supervision as well as an execution mechanism.

The study uses fixed fixtures and limited evaluation sets. Transfer across robots and end effectors, and robustness to disturbances, workpiece motion, friction, and mating-clearance changes, remain to be evaluated under controlled variations.

\FloatBarrier
\appendices
\section{Physical Interpretation of Interaction Costs and Policy Evaluation}
\label{app:physical_analysis}
For continuous signals on a fixed interval $\mathcal I_k$, increasingly fine sampling makes $\Delta t\sum_jP_c(t_{k,j})$ approach $\int_{\mathcal I_k}P_c(t)\,dt$, and $\max_jq(t_{k,j})$ approach $\sup_{t\in\mathcal I_k}q(t)$, where $q$ is the gated, capped term in \eqref{eq:tail}. The costs thus preserve conflict duration and the strongest directional high-force event, respectively.

Let $n_k(P_0)$ count samples with $P^F_{c,k,j}\ge P_0>0$, and let $\mathcal E_k$ denote the existence of a sample with $\|\widetilde F^{\rm raw}_{k,j}\|\ge F_0>F_{\rm soft}$ and $P^{\rm raw}_{L,k,j}\ge P_0$. Set $g_0=g^F(F_0)$ and $q_0=\min(P_0/P_F^{\rm ref},q_{\max})$. Nonnegativity of rotational conflict and monotonicity of the force gate give
\begin{equation}
n_k(P_0)\le\frac{E_F^{\rm ref}c_{c,k}}{\Delta tP_0},\qquad
\mathbf1_{\mathcal E_k}\le\frac{c_{t,k}}{g_0q_0}.
\label{eq:durationbound}
\end{equation}
For $C_i(\pi)=\mathbb E_\pi[\sum_{k\ge0}\gamma^kc_{i,k}]$, $i\in\{c,t\}$, discounted summation and expectation yield
\begin{align}
\mathbb E_\pi\!\left[\sum_k\gamma^kn_k(P_0)\right]&\le\frac{E_F^{\rm ref}C_c(\pi)}{\Delta tP_0},\label{eq:trajectorydurationbound}\\
\mathbb E_\pi\!\left[\sum_k\gamma^k\mathbf1_{\mathcal E_k}\right]&\le\frac{C_t(\pi)}{g_0q_0}.
\label{eq:trajectoryeventbound}
\end{align}
Equation~\eqref{eq:durationbound} follows because each counted conflict sample contributes at least $\Delta tP_0/E_F^{\rm ref}$, while an event in $\mathcal E_k$ makes the tail maximum at least $g_0q_0$. Discounted summation preserves both inequalities. The resulting bounds describe expected conflict exposure and high-force continued-loading events; withdrawal and force-orthogonal corrections remain unpenalized by the directional term.

In the same fixed-admittance MDP and policy class, define $R(\pi)=\lambda_cC_c(\pi)+\eta_FC_t(\pi)$. If $\pi_0$ and $\pi_*$ globally maximize $J_{\rm task}$ and $J_{\rm task}-R$, respectively, their optimality inequalities give
\begin{equation}
R(\pi_*)\le R(\pi_0)+J_{\rm task}(\pi_*)-J_{\rm task}(\pi_0)
\le R(\pi_0).
\label{eq:weightedrisk}
\end{equation}
Here $J_{\rm task}$ includes the common time penalty. This ideal result bounds the weighted cost, not each component separately.

Finally, we examine whether fixed-policy value evaluation retains a unique solution after adding the interaction costs. The tail contribution is bounded by $\eta_Fq_{\max}$. If reachable wrench, commanded twist, and policy-window duration are bounded, the conflict term and total reward are also bounded. On an augmented Markov state or equivalent sufficient history, the fixed-policy evaluation operator is
\begin{equation}
(\mathcal T^\pi Q)(s,a)=r(s,a)+\gamma
\mathbb E_{s',a'\sim\pi}[m(s')Q(s',a')],
\end{equation}
where $m\in[0,1]$ is the terminal mask. For bounded rewards, it satisfies
\begin{equation}
\|\mathcal T^\pi Q_1-\mathcal T^\pi Q_2\|_\infty
\le\gamma\|Q_1-Q_2\|_\infty.
\end{equation}
The inequality shows that the fixed-policy Bellman operator remains a $\gamma$-contraction after adding bounded interaction costs. Thus, with $\gamma=0.97<1$, exact fixed-policy Bellman iteration converges to the unique policy value function. Reward shaping therefore preserves this basic property of fixed-policy evaluation.

\bibliographystyle{IEEEtran}
\bibliography{references}
\end{document}